\documentclass[preprint,12pt]{elsarticle}

\usepackage{amssymb}
\usepackage{times}
\usepackage{graphicx}
\usepackage{algorithm}
\usepackage{amsmath}
\usepackage[noend]{algorithmic}
\usepackage{bm}
\usepackage{latexsym}
\usepackage{booktabs}
\journal{Nuclear Physics B}

\begin{document}

\begin{frontmatter}



\title{Dynamic Transformer Architecture for Continual Learning of Multimodal Tasks}


\author{Yuliang Cai and Mohammad Rostami}

\affiliation{organization=University of Southern California\\%Department and Organization
            Los Angeles, CA, USA.}

\begin{abstract}
Transformer neural networks are increasingly replacing prior architectures in a wide range of applications in different data modalities. The increasing size and computational demands of fine-tuning large pre-trained transformer neural networks pose significant challenges for the widespread adoption of these models for applications that demand on-edge computing. To tackle this challenge, continual learning (CL) emerges as a solution by facilitating the transfer of knowledge across tasks that arrive sequentially for an autonomously learning agent. However, current CL methods mainly focus on learning tasks that are exclusively vision-based or language-based. We propose a transformer-based CL framework focusing on learning tasks that involve both vision and language, known as Vision-and-Language (VaL) tasks. Due to the success of transformers in other modalities, our architecture has the potential to be used in multimodal learning settings. In our framework, we benefit from introducing extra parameters to a base transformer to specialize the network for each task. As a result, we enable dynamic model expansion to learn several tasks in a sequence. We also use knowledge distillation to benefit from relevant past experiences to learn the current task more efficiently. Our proposed method, Task Attentive Multimodal Continual Learning (TAM-CL), allows for the exchange of information between tasks while mitigating the problem of catastrophic forgetting. Notably, our approach is scalable, incurring minimal memory and time overhead. TAM-CL achieves state-of-the-art (SOTA) performance on challenging multimodal tasks\footnote{Early partial results of this work are presented in EMNLP 2023~\cite{cai2023task}}.
\end{abstract}


% \begin{highlights}
% \end{highlights}


\end{frontmatter}


\section{Introduction}

Large pre-trained transformer models have found application in a wide range of domains, encompassing tasks that involve both vision and language ~\cite{dosovitskiyimage,kim2021vilt,xu2023multimodal,li2023lvit,zhang2023vitaev2,ali2023vision}.
The transformer architecture is based on encoder and decoder layers that implement the self-attention mechanism~\cite{vaswani2017attention}. 
The self-attention layers  implement an attention mechanism on an input sequence by relating the positions of tokens in the sequence to the way that a global feature vector is extracted from the sequence. 
 The idea of self-attention then was adopted for other modalities such as computer vision tasks~\cite{dosovitskiy2020image,han2022survey} and speech processing tasks~\cite{karita2019comparative,wang2020transformer} based on modeling an image using a sequence of  patches as tokens.  Various improved transformer architectures are developed by modifying the base idea and training procedures for each of the language and image modalities.

Typically, transformer models undergo pretraining on a substantially large dataset, followed by fine-tuning to adapt to a specific downstream task, even with limited task-specific data. 
This approach enables effective transfer learning   to leverage the generalizable knowledge gained from the large dataset. 
However, fine-tuning at the task level can weaken the model's ability to generalize because the model learns to become task-specific. It also requires the retention of a separate version of the base model for each task. As transformer architectures are persistently becoming larger and larger, storing many versions of a base architecture is becoming a practically infeasible approach. To address these issues, continual learning (CL) ~\cite{jin2021learn,yang2022continual,wang2022continual,pelosin2022towards,ermis2022continual,srinivasan2023i2i,mohamed2023d3former,liu2023continual,wang2023continual} algorithms have been developed to navigate these complexities in transformer models. They employ a shared model that leverages cross-task knowledge transfer to learn tasks that are encountered sequentially.  The goal is to enable the shared model to learn new tasks without degraded performance on the previously learned tasks.

Continual learning has a rich precedent before the invention of transformers~\cite{silver2013lifelong,chen2018lifelong,rostami2017multi,rostami2020using,van2019three}.
The primary obstacle in Continual Learning (CL) is known as catastrophic forgetting ~\cite{french1999catastrophic}, where the model performance on past learned tasks degrades as the model is updated to learn new tasks. There are several approaches to address catastrophic forgetting. One category of CL algorithms tackles this challenge by regularizing a fixed shared model to learn different tasks through distinct information pathways, or weights ~\cite{kirkpatrick2017overcoming,aljundi2018memory,han2023online}. The fundamental concept of regularization-based methods is to identify a subset of model parameters crucial for encoding the acquired knowledge of each task and then consolidate these parameters when updating the model for new tasks. Another approach involves model expansion ~\cite{rusu2016progressive,yoonlifelong}. The goal is to enlarge a base model with a small set of additional weights and applying the network to learn new tasks using these weights. Lastly, some algorithms employ pseudo-rehearsal with experience replay ~\cite{robins1995catastrophic,rostami2019complementary,rolnick2019experience,atkinson2021pseudo,rostami2021lifelong,mirtaheri2023history}. This approach entails storing a representative sample of training data for each task in a memory buffer and replaying them alongside the current task's data to retain the encoded knowledge of past tasks. Certain methods alleviate the need for a memory buffer by enabling the model to generate pseudo-samples for previously learned tasks, which are then employed in experience replay. Despite their effectiveness, existing CL techniques are designed for unimodal tasks, such as tasks solely involving vision ~\cite{rostami2020generative,lin2021clear,https://doi.org/10.48550/arxiv.2111.11326,hu2023class} or language ~\cite{jin2021learn,yang2022continual}, and do not address the distinct challenges posed by multimodal tasks, such as Vision-and-Language (VaL) tasks. A major reason behind unimodal approach is that prior to the transformer architectures, the specific neural network architectures for each modality were quite different, e.g., convolutional neural networks vs recurrent neural networks. Because the base idea of transformers work well in different data modalities, we may be able to address multimodal learning with a single base unified architecture to fuse inputs from different modalities.

More specifically, we introduce a novel algorithm for learning tasks involving both vision and language (VaL) within a CL framework. In our work, we utilize dynamic model expansion in the context of transformers. To achieve this end, we utilize the self-attention layers of a foundational bimodal transformer as a shared encoder across all tasks. Subsequently, we enhance the base model with task-attention layers \cite{https://doi.org/10.48550/arxiv.2111.11326,azad2022contextual}, which serve to customize the model for each task by incorporating a task-specific token. Our approach entails a minimal memory overhead and introduces only a slight increase in inference time during testing. Furthermore, it operates without the need for extensive hyper-parameter tuning and remains resilient even in scenarios where the number of tasks is unknown. Our distinct contributions include:

\begin{itemize}
    \item  A dynamically expanding, efficient transformer architecture for multimodal CL based on task attention layers that make the model task attentive. 
    \item  A training algorithm to handle diverse, sequentially arriving vision-and-language  tasks such as visual question answering, visual entailment, and visual reasoning.
    \item  Extensive experiments to demonstrate that the proposed model achieves SOTA performance compared to existing baselines.
\end{itemize}

\section{Background and Related Work}

\paragraph{Transformers for Vision and Language Tasks}
Multimodal transformers have been developed for processing tasks that involve both vision and language (VaL) ~\cite{su2019vl,tan2019lxmert,kim2021vilt,viltbert,chen2020uniter}. The fundamental concept behind these transformers is to utilize self-attention layers on each modality to extract relevant features from both the visual and linguistic inputs and fuse the knowledge to solve the task with more comprehensive understanding provided by multi-modalities. As knowledge fusion is the key component of multimodal transformer, several different strategies has been applied to boost the performance, such as early summation \cite{gavrilyuk2020actortransformers, xu2022deepchange}, early concatenation \cite{sun2019videobert,shi2022learning}, and cross attention \cite{murahari2020largescale} that fuse the knowledge by different methods and different stage of feature processing. The features are subsequently fused at higher layers to derive cross-modal contextualized representations of the multimodal inputs. The objective is to capture the interaction between the visual and linguistic inputs, which is crucial for effectively performing VaL tasks. This approach has also been applied in other domains, such as in the context of processing video to establish relationships between visual and speech inputs ~\cite{arnab2021vivit,sun2019videobert}. Multimodal transformers are typically trained on extensive datasets and have demonstrated remarkable effectiveness when fine-tuned for specific tasks downstream. Given the substantial performance gains they offer, transformers are progressively supplanting older architectural paradigms.

\paragraph{Continual Learning}
The primary challenge in continual learning is addressing catastrophic forgetting~\cite{french1999catastrophic}. A dominant approach to address this challenge is  experience replay~\cite{schaul2015prioritized,zhang2017deeper,fedus2020revisiting,rostami2023overcoming,rostami2021detection,shin2017continual,rostami2021cognitively,stan2021unsupervised}. The main idea for experience replay is  to store and then replay representative samples of past tasks alongside the current task data to preserve the learned distributions through pseudo-rehearsal.  
 Weight consolidation using structural plasticity~\cite{kirkpatrick2017overcoming,kj2020meta,cong2023self} is another strategy to approximate experience replay. The concept is to identify important weights that retain knowledge about a task and then consolidate them based on their relative importance for past tasks in the future~\cite{kirkpatrick2017overcoming} and learn new tasks through weights that are unimportant for past learned tasks.
These approaches are not efficient for transformers due to the large size of these models. A more practical approach is based on the notion of adapters \cite{ke2021achieving,ermis2022memory,ermis2022continual}, where most transformer weights remain frozen and a small number of trainable weights are used to make the model dynamic.
In this context, mechanisms such as experience replay or weight consolidation can be used on the adapter weights to overcome forgetting effects.

\paragraph{Knowledge Distillation for Continual Learning}
Despite the impressive success of deep learning, neural networks models  fail to be deployed in limited-sourced devices such as phone and embedded devices due to the significant computational load of their training. Thus, the compression and acceleration of the model is urgently needed. 
First introduce by Bucilua et al \cite{10.1145/1150402.1150464}, knowledge distillation is widely used to compress the knowledge of a large and complex model, the teacher model, into a smaller and efficient ones, the student models. Fitnet \cite{romero2015fitnets} deploys the knowledge distillation between the intermediate representations between teacher and student model to improve the training of the students. On the other hand, Chen et al. \cite{Chen2017LearningEO} utilize the output, $\textit{logit z}$, of student and teacher model to compute the knowledge distillation loss.
Lopez-Paz and Ranzato \cite{lopezpaz2022gradient} developed one of the first methods to adopt knowledge distillation in to the continual learning setting. Within the scenario of learning from $\mathcal{T}$ sequential tasks, the copied learnt model on the previous task, $t_{i-1}$, is used as the teacher model to guide the training of the model on the current task $t_i$. The knowledge distilled from the previous model can efficiently help preserve the previous knowledge and mitigate catastrophic forgetting, while at the same time accelerate the forward transfer of the knowledge to learn the current task efficiently.

\paragraph{Continual Learning for Transformer}
  Fine-tuning transformers inherently sacrifices their ability to generalize effectively. Employing a separate transformer for each task would result in a substantial increase in disk memory usage, given the significant growth in transformer sizes to handle more complex tasks. While CL appears to be a natural remedy for these challenges, research on CL with transformers remains relatively scarce.
Xin et al.~\cite{jin2021learn} utilize adapters in conjunction with a hypernetwork~\cite{von2019continual} to enable CL for language tasks. Alternatively, Yang et al.~\cite{yang2022continual} propose a transformer calibration module to enhance a transformer's adaptability. This module operates independently from the base pre-trained transformer and aids in adopting it for a specific downstream task.
The Lifelong Vision Transformer \cite{wang2022continual} integrates information from previous tasks through an inter-task attention mechanism, thus reducing the rate at which critical attention weights shift away from older tasks towards the current one. Douillard et al. ~\cite{https://doi.org/10.48550/arxiv.2111.11326} propose a CL architecture for vision tasks using the ViLT model ~\cite{kim2021vilt}. Pelosin et al.~\cite{pelosin2022towards} extend this approach to an exemplar-free setting by distilling the attention-level matrices of transformers. This enables model adaptability and helps mitigate the effects of forgetting. Ermis et al. ~\cite{ermis2022continual} apply the concept of adapters in a vision context.
To the best of our knowledge, there has been no prior exploration of CL for multimodal tasks using transformer architectures.

\section{Problem Description}

\begin{figure}[t!]
\centering
\includegraphics[scale=.8]{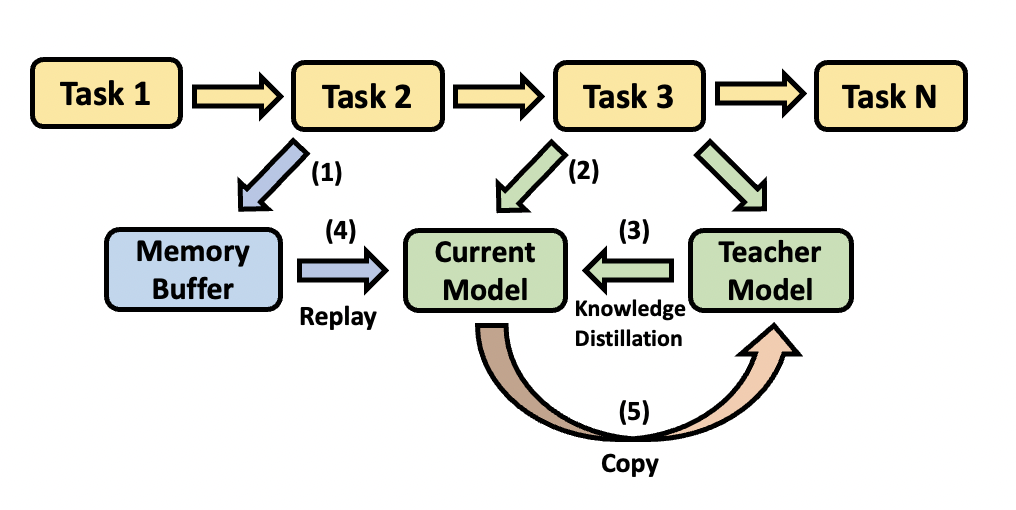}
\caption{\small The proposed  CL training procedure: (1) A small portion of the data for previous tasks are randomly selected and stored in a memory buffer. (2) The current task  arrives with $\mathcal{D}^i$. (3) The training data $\mathcal{D}^i$ is used  as input to the teacher model to compute the   distillation loss. (4) The memory buffer samples are replayed along with the current task data to train the main model. (5) After learning the current task,   the teacher model of the next task will be a copy of the current   model.}
\vspace{-3mm}
\label{fig:Diagram}
\end{figure}

Consider a collection of VaL tasks $\{\mathcal{T}_i\}_{i=1}^T$ which are introduced in a sequential manner, and each task $\mathcal{T}_i$ comes with an annotated training dataset $\mathcal{D}^i$ = $\{\langle(\bm{I}_i^j,\bm{L}_i^j)^i, y^j_i\rangle_{j=1}^{N_i}\}$, where $\bm{I}_i^j \in \mathbb{R}^{H\times W\times C}$ represents the image input,  $\bm{L}_i^j \in \mathbb{R}^{L\times |V|}$ represents the language input, while $y^j_i$ is the text-typed discrete label. It's worth noticing that the order of these tasks and $T$ are not known in advance and the agent encounters the task without any advance knowledge. The training data points for $\mathcal{T}_i$ are assumed to be drawn iid from a task-specific joint distribution $p_i^t(\cdot,\cdot,\cdot)$. 
In the context of multimodal continual learning, our objective is to learn each task at time-step $i$ and then progress to learn the next tasks. 
The learned tasks can be encountered at any time during testing in the future. As a result, we would like to preserve the performance achieved on the previous learned tasks through the prevention of catastrophic forgetting.

When learned individually, each of these VaL tasks $\mathcal{T}_i$ can be learned using supervised learning conditioned on selecting the suitable predictive model $f^i_{\theta_M}(\cdot,\cdot)$, e.g., a transformer  with trainable parameters $\theta_M$, and the discrimination loss   $\mathcal{L}(\cdot)$, e.g., cross-entropy. 
However, given storage constrains, we assume a shared model must be employed, as training a separate model per task is impractical. Moreover, only a limited portion of training data for each task $\mathcal{T}_i$ can be retained post-training because normally the training data is not stored due to storage limitations, which makes multitask learning ~\cite{caruana1998multitask,hu2021unit,bhattacharjee2022mult} an unfeasible solution. 
The single task learning strategy using a shared model is not ideal in CL either. Because when the model is updated to learn the current tasks, its performance on the past   tasks may suffer from catastrophic forgetting due to updates in the model parameters, leading to a degradation in performance.
On the other hand, single task learning does not benefit from knowledge transfer because the tasks are learning   in isolation.

Figure \ref{fig:Diagram} provides an overview of our proposed solution to address multimodal CL.
To leverage a shared model across all tasks and facilitate knowledge transfer across all tasks, we utilize a base transformer model as  $f_\theta(\cdot,\cdot)$. The base model remains largely frozen during the training phase to help maintaining its generalization power. The model   is also made adaptive by incorporating a unique task attention layer after its final layer. Additionally, we modify the   supervised learning loss by introducing a knowledge distillation loss ~\cite{hinton2015distilling} on the intermediate model layers to enable positive forward knowledge transfer. We consider the teacher model in the knowledge distillation formulation to be a copy of $f^{i-1}_{\theta}(\cdot,\cdot)$ when training the main student model on $i^{th}$ task $\mathcal{T}_i$.  The teacher model is served as an intermediate means of transferring past experiences. Additionally, we employ pseudo-rehearsal through experience replay \cite{rolnick2019experience}, utilizing a small memory buffer to mitigate the effect of catastrophic forgetting ~\cite{kirkpatrick2017overcoming}.   

\section{The Proposed Architecture}

\begin{figure*}
\centering
\includegraphics[scale=0.44]{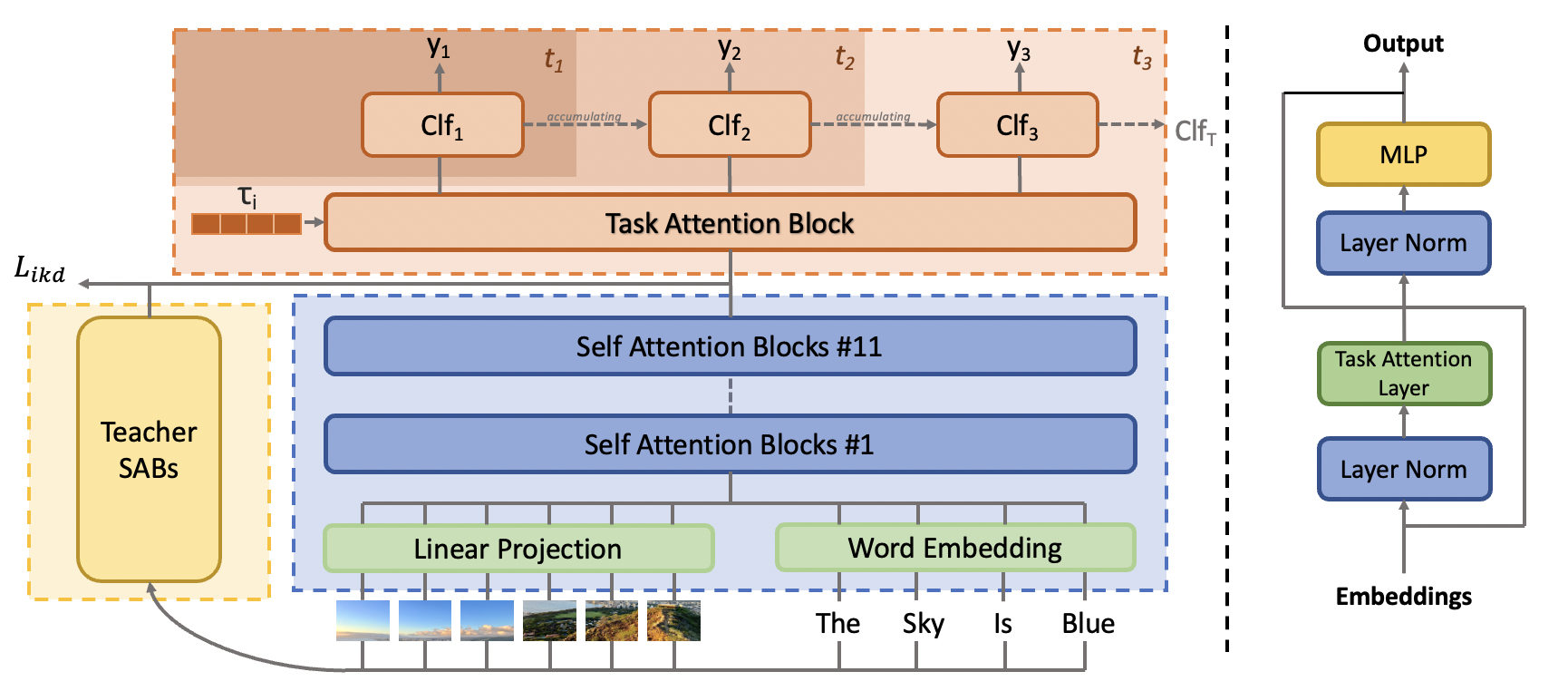}
\caption{The proposed transformer-based architecture: (left) The VaL inputs are converted into two sequences and then fed into the self-attention layers to generate a fused global feature vector. The data feature vector is then concatenated with the learnable task-specific tokens and then fed into the task attention layer to generate the input for the task-specific classifier heads. The same VaL inputs are also fed into the teacher model's transformer architecture to compute Knowledge Distillation. (right) The task-attention block architecture.}
\label{fig:Big_Diagram}
\end{figure*}

More specifically, Figure~\ref{fig:Big_Diagram} visualizes our transformer-based architecture for continual learning in multimodal learning scenario. The architecture is composed of a shared pre-trained replaceable multimodal transformer, a shared non-pretrained task attention block, and MLP classification headers that are task-specific. Using task-specific heads allow for learning tasks that do not share the same output-space. To make the model adaptive to sequentially coming tasks, The task attention block receives task-specific tokens, estimated during learning each task, task-specific tokens are initialized and sent to task attention blocks along with the input instances to generate the expected output. We provide details about these modules and the strategies that we use to train it.

\subsection{Sequential Feature Extraction Block}
\label{4.1}
We refer to our architecture as Task Attentive Multimodal Continual Learning (TAM-CL). In TAM-CL, we adopt the ViLT \cite{kim2021vilt} feature generation procedure to encode the visual inputs. To this end, we first break down a given image $\bm{I} \in \mathbb{R}^{H\times W \times C}$ into a sequence of patches. These patches are then flattened to create 2D vectors $\bm{v} \in \mathbb{R}^{N\times (P^2 \cdot C)}$. In this context,$C$ denotes the number of channels,
$P\times P$ represents the size of each image patch, and $N = HW/P^2$ stands for the number of patches. Next, we apply a trainable linear projection represented by $\bm{V} \in \mathbb{R}^{(P^2 \cdot C) \times H}$. These transformations convert $\bm{v}$ into a sequential representation $\bm{\overline{v} }\in \mathbb{R}^{N\times H}$ according to the equation:
\begin{equation}%\small
    \bm{\overline{v}} = [\bm{v}^{\text{class}};\bm{v}_1\bm{V};\bm{v}_2\bm{V};...;\bm{v}_N\bm{V}] + \bm{V}^{\text{pos}}.
\end{equation}

Here, $\bm{v}^{class}$ corresponds to class-specific vectors, and $\bm{V}^{pos}$ stands for the position embedding. This process enables the creation of a sequential representation of the image patches, which can be subsequently used in the following layers of the model for further processing and abstract feature extraction.

We also derive word vectors for $\bm{l} \in \mathbb{R}^{L\times |V| }$ to encode the language input. This task is achieved by applying a word embedding matrix $\bm{T} \in \mathbb{R}^{|V|\times H}$ and incorporating a position embedding matrix $\bm{T}^{pos} \in \mathbb{R}^{(L+1)\times H}$. This process results in the embedded data $\bm{\overline{t}} \in \mathbb{R}^{L\times H}$ according to the equation:
\begin{equation}%\small
    \bm{\overline{t}} = [\bm{t}^{\text{class}};\bm{l}_1\bm{T};\bm{l}_2\bm{T};...;\bm{l}_L\bm{T}] + \bm{T}^{\text{pos}}.
\end{equation}
Here, $\bm{t}^{\text{class}}$ pertains to class-specific vectors, and $\bm{T}^{\text{pos}}$ denotes the position embedding. Subsequently, we add the image and text embeddings to their respective independent model-type embedding vectors, namely $\bm{t}^{type}$ and $\bm{v}^{type} \in \mathbb{R}^H$. These components are then concatenated to form a unified sequence $\bm{s}$ as follows:
\begin{equation}%\small
    \bm{s^0} = [\bm{v^{\text{type}}}+\bm{\overline{v}};\bm{t^{\text{type}}}+\bm{\overline{t}}]
\end{equation}

The amalgamated vector $\bm{s}$ is processed through $D$ standard self-attention layers within the base VaL transformer (ViLT). The output embeddings from these attention layers are referred to as $\bm{s}^D$, and they are subsequently passed into the task attention layer. The resulting embeddings from the   attention layers are denoted as $\bm{s}^{D+1}$.
The transformations are carried out in two steps, described as:

\begin{equation}%\small
\begin{aligned}
    &\bm{\hat{s}}^{d} = MSA(LN(\bm{s}^{d-1}))+\bm{s}^{d-1},    d = 1,...,D \\
    &\bm{s}^{d} = MLP(LN(\bm{\hat{s}}^{d}))+\bm{\hat{s}}^{d},   d = 1,...,D
\end{aligned}
\end{equation}

Despite fusing the multimodal inputs, the model would be a single-task learning model. To enable learning multiple tasks, we need a task-dependent block.

\subsection{Task Attention Block}
\label{section4.2}
The essence of our approach hinges on adopting task-level self-attention each time a new task is learned by introducing task-specific tokens. The notion of task token has been used before in classic CL to make a base transformer model adaptive~\cite{isele2016using}. In contrast to the standard self-attention layer, the task-attention layer is designed specifically-associated with the current task. For each new task $i$ within the range of $1$ to $T$ comes in, a trainable task token $\tau_i \in \mathbb{R}^{G\times 1}$ is initialized and applied as the input of the task-attention layer for the current task, where $G$ represents the size of the latent space for each self-attention layer. During the training step for each task, the task token is estimated to generate the desired output for that task. Much like the self-attention block (SAB), the task-attention block (TAB) is a module comprising an attention layer, layer normalization, and a Multi-Layer Perceptron (MLP). However, in this case, the attention layer is specifically designed as a task-attention layer, deviating from the standard self-attention, which is task-specific, by using modified queries.

For task $i$, the task attention block receives two inputs: the batched output from the previous self-attention blocks $\bm{s}_{ib}^D$, where $b$ notes the specific batch from the training data of $T_i$, and the task-specific token $\tau_i$. Notably, the same task token $\tau$ is employed for all instances in the task-specific training/testing dataset. These two vectors are concatenated to form an input for the task attention:

\begin{equation}%\small
    \begin{aligned}
    &\bm{s}{'}_i^{D+1} = [\tau_i, \bm{s}_i^{D}] \in \mathbb{R}^{(N+1)\times G}, i = 1,...,T\\
        &\bm{\hat{s}}_i^{D+1} = \textit{TA}(LN(\bm{s}{'}_i^{D+1})), i = 1,...,T \\
&\bm{s}_i^{D+1} = \textit{MLP}(LN(\bm{\hat{s}}_i^{D+1})) + \bm{\hat{s}}_i^{D+1}, i = 1,...,T\\
    \end{aligned}
\end{equation}

Here, $LN$ denotes layer normalization, and $TA$ represents the task-attention layer. This process is performed for each task within the range of $1$ to $T$.

The task attention block is added after the final self-attention block of the backbone transformer. While it's possible to incorporate more than one task attention block, our architecture utilizes a single TAB for simplicity. The functioning of the task attention layer is described as follows:

\begin{equation}%\small
\begin{aligned}
&\bm{Q}_i = \bm{W}_q \times \tau_i, \\
&\bm{K}_i = \bm{W}_k \times \bm{s}_i^{D+1}, \\
&\bm{V}_i = \bm{W}_v \times \bm{s}_i^{D+1}, \\
&\bm{A}_i = Softmax(\bm{Q}_i \cdot \bm{K}_i^T/\sqrt{G/h}), \\
&\bm{O}_i = \bm{W}_o\bm{A}_i\bm{V}_i + \bm{b}_o \in \mathbb{R}^{1 \times G},
\end{aligned}
\end{equation}
where $h$ represents the number of attention heads in the transformer \cite{https://doi.org/10.48550/arxiv.1706.03762}. Additionally, $\bm{W}_q$, $\bm{W}_k$, $\bm{W}_v$ and $\bm{W}_o$ are trainable weight matrices, while $\bm{b}_o$ is a bias vector.
Subsequently, the output of the task attention block, $\bm{s}_i^{D+1}$, is fed into task-specific classifier layers:

\begin{equation}%\small
    \bm{y}_i = \textit{Clf}_i(\bm{s}_i^{D+1}), i = 1,...,T,
\end{equation}
where  $\textit{Clf}_i$ represents the classifier subnetwork, which is far smaller than  prior blocks in size, specific to task $i$ to generate the final output  for task-specific predictions. Although the model becomes adaptive with this block, catastrophic forgetting remains a challenge if we use this model to learn sequential tasks. To address this challenge we relay on knowledge distillation and experience replay.

\subsection{Teacher Model Block}

Besides the architecture proposed in section \ref{4.1} and \ref{section4.2}, to further consolidate the weights that maintains the knowledge of previous tasks, the teacher model is applied parallel to the main model, noted student model, such that the same input data can be sent to both the teacher model and student model for the computation of knowledge distillation loss to enable forward positive knowledge transfer.

Consider $\mathcal{V}$ as the base transformer ViLT mentioned in section \ref{4.1}, and $\mathcal{V}_i$ as model $\mathcal{V}$ after trained on $i^{th}$ task. Denote $\mathcal{M}_s$ as student model and $\mathcal{M}_t$ as teacher model. During the training on task $i$, we have:

\begin{equation}%\small
\begin{aligned}
& \mathcal{M}_s = \mathcal{V}_i, \\
& \mathcal{M}_t = \mathcal{V}_{i-1}.
\end{aligned}
\end{equation}
where $\mathcal{M}_t$ is frozen completely. The detail of our approach knowledge distillation will be discussed in Section \ref{section: 5.2}.

\section{Training Algorithm}
Note that the architecture depicted in Figure~\ref{fig:Big_Diagram} provides an overview of our model, TAM-CL, at a specific time step. This architecture dynamically expands  as additional tasks are learned in succession and more task tokens are accumulated. However, the addition includes the classification heads and tokens which makes it significantly more efficient compared to learning independent models, one per task. We will now elaborate on the appropriate CL loss functions that we employ for the training process the model and the task tokens.

\subsection{Token Expansion}
During the training phase, the self-attention blocks of the base transformer, as well as the task attention block, are shared across all tasks. However, for each new task, we introduce a new task token, denoted as $\tau \in G\times 1$, with the same dimension and initialize it uniformly. The token is then updated to fit well with the new task. Additionally, we initialize a new cumulative task-specific classifier, $\textit{Clf}_i(\cdot)$, for task $i$. The output dimension of $\textit{Clf}_i(\cdot)$ is expanded based on the previous classifier, $\textit{Clf}_{i-1}(\cdot)$, to include new outputs for the current task. As more tasks are added, the output dimension of the task-specific classifier $i$ accumulates as follows:

\begin{equation}%\small
    \bm{E}_i = \bm{E}^{\textit{orig}}_i + \bm{E}_{i-1}, i = 1,\ldots,T.
\end{equation}

Here, $\bm{E}_i$ represents the output dimension for the $i$-th classifier, while $\bm{E}^{\textit{orig}}_i$ signifies the output dimension originally designed for the $i$-th task.

For the $i$-th task, we combine the $i$-th task token with the updated path token $\bm{s}^D$ from the last self-attention block of the transformer, and input it into the task attention block, as detailed in Section 4.2. During this stage, only the $i$-th task token and task-specific classifier are trainable, while all other task tokens and classifiers remain fixed. Since the classifier heads are significantly smaller than the base network, storing them do not lead to a significant memory overhead.

In the testing phase, the task number $i$ of the test data is explicitly provided. $\bm{s}^D$ is combined with the $i$-th learned task token and fed into the task attention block, along with utilizing its corresponding task-specific classifier. All other task tokens and classifiers remain inactive.

\subsection{Loss and Knowledge Distillation}
\label{section: 5.2}
The objective function in TAM-CL consists of three distinct loss components. First, the cross-entropy loss, $\mathcal{L}_c$, serves as the original task-specific loss function used in single-task learning, with its value varying depending on the specific task at hand. Second, the knowledge distillation (KD) loss, $\mathcal{L}_{ikd}$, is computed based on the output of the last self-attention block $\bm{s}^D$, from both the main student model and the teacher model. This loss term is crucial for transferring knowledge from the teacher to the student model, effectively constraining distribution shifts and preventing catastrophic forgetting as well. Finally, the diverse loss, $\mathcal{L}_{div}$ evaluates the diversity of the data distribution associated with task tokens, encouraging greater diversity among them to enable learning task-specific knowledge and mitigates the effect of negative knowledge transfer~\cite{zhang2022survey}.
Our final loss function is a weighted combination of these components:
\begin{equation}%\small
    \mathcal{L} = (1-\lambda)\mathcal{L}_c + \lambda\alpha\mathcal{L}_{ikd} + \beta\mathcal{L}_{div},
\end{equation}
where $\lambda$ is determined as $\frac{T_n-1}{T_n}$ and $T_n$ represents the total number of tasks encountered thus far. The parameter $\alpha$ is task-specific and $\beta$ is computed as $\beta = min(\mathcal{L}_{div}, 0.1\times ((1-\lambda)\mathcal{L}_c + \lambda\alpha\mathcal{L}_{ikd}))$.
The intermediate knowledge distillation, $\mathcal{L}_{ikd}$, is a central aspect of TAM-CL. It enables the transfer of knowledge from the teacher model to the main student model, effectively limiting the probability shift of pre-trained parameters. This, in turn, allows subsequent layers to remain flexible in adapting to new tasks. Notably, TAM-CL introduces this intermediate knowledge distillation objective function in the context of multimodal continual learning, marking a significant advancement in the field.

\subsection{Experience Replay}
The described training procedure above allows for the training of a shared model across tasks, although it necessitates datasets for all tasks, which challenges a fundamental assumption in CL. Because when a task is learned, we may not have access to its data when learning future tasks. On the other hand, catastrophic forgetting is a general issue if we exclude the full dataset for the previously learned tasks. To address these issues, we rely on the classic experience replay approach. We employ a memory buffer during the training phase at each time-step. This buffer retains a small percentage of randomly selected samples, e.g., approximately 5\% in our experiments, of the training dataset for all previous tasks. When learning the current task with a specific batch number, the subsequent batch is chosen at random from the memory buffer and is used to represent the past tasks. This process helps consolidate the parameter distribution related to previous tasks, effectively mitigating forgetting effects.
We present the Task Attentive Multimodal Continual Learning (TAM-CL) training procedure in to Algorithms 1 and 2.

\begin{algorithm}[t]
%\small
\caption{TAM-CL Train}
\begin{algorithmic}
\STATE \textbf{INPUT:} Model \textbf{M}, MemBuffer \textbf{B}, ReplayFreq \textbf{f}\\
\FOR{epoch in num\_epoch}
\FOR{step, batch in dataloader}
\STATE Loss $\leftarrow$ TrainStep(\textbf{M}, batch)
\IF{step \% \textbf{f} == 0}
\STATE $\text{batch}_{replay}$ $\leftarrow$ \textbf{B}.getBatch()
\STATE $\text{Loss}_{replay}$ $\leftarrow$
TrainStep(\textbf{M}, $\text{batch}_{replay}$)
\ENDIF
\ENDFOR
\ENDFOR
\end{algorithmic}
\end{algorithm}

\begin{algorithm}[t]
%\small
\caption{TAM-CL TrainStep}\label{alg:step}
\begin{algorithmic}
\STATE\textbf{INPUT:} Model \textbf{M}, Teacher Model \textbf{T}, Batch, Target \textbf{t}, Token \textbf{k}
\STATE p $\leftarrow$ Model(Batch)
\STATE $loss$ $\leftarrow$ CrossEntropyLoss(p, \textbf{t})
\STATE ModelSabPre $\leftarrow$ \textbf{M}.SAB(Batch)
\STATE TeacherSabPre $\leftarrow$ \textbf{T}.SABs(Batch)
\STATE $loss_{ikd} \leftarrow$ KL-Div(ModelSabPre, TeacherSabPre)
\STATE $loss_{div} \leftarrow$ CrossEntropyLoss($\textbf{k}_i,\textbf{k}_j)\quad j=1,..,i-1$\\
$loss$ $\leftarrow$ $(1-\lambda)loss$ + $\lambda\alpha loss_{ikd}$ + $\beta loss_{div}$\\
 $loss$.backward()\\
\textbf{Return} loss
\end{algorithmic}
\end{algorithm}

\section{Experiments Result}

As a method proposed to solve vision-language tasks, TAM-CL needs to be evaluated using tasks   such as visual question answering, visual understanding and visual reasoning. To test its ability to prevent   catastrophic forgetting, we choose   five independent multimodal datasets for evaluation which are presented to the architecture in a sequence. Additionally, to show that TAM-CL is a competitive method, we choose five  baseline methods for comparisons. These methods include  fine-tuning as a lowerbound and four continual learning methods.

\subsection{Experiments Setup}
\label{sec:6.1}

\begin{table}
%\small
\centering
\begin{tabular}{c|c|c}
    \toprule
    Name & \# Training Examples & \# Labels\\ \midrule
    NLVR2 & $80000$ & $2$\\
    SNLI-VE & $80000$ & $3$\\ 
    COCOQA &  $78736$ & $430$\\ 
    GQA & $80000$ & $1842$\\ 
    OKVQA & $18032$ & $2910$ \\
    \bottomrule
\end{tabular}
\caption{Statistics of the VaL datasets used in the experiments.}
\label{tab:dataset}
\end{table}

\paragraph{Datasets} we use \textbf{SNLI-VE}~\cite{https://doi.org/10.48550/arxiv.1901.06706}, an image-sentence pairs dataset whereby a premise is defined by an image, rather than a natural language sentence as in traditional Textual Entailment tasks, \textbf{COCOQA}~\cite{https://doi.org/10.48550/arxiv.1505.02074}, a visual question answering dataset based on Microsoft COCO image dataset, \textbf{GQA}~\cite{hudson2019gqa}, a compositional question-answering and visual reasoning dataset leverages scene graph structure, \textbf{NLVR2}~\cite{https://doi.org/10.48550/arxiv.1811.00491}, a visual reasoning dataset which takes two images and determine the correctness of the given sentence, \textbf{OKVQA}~\cite{marino2019okvqa}, a knowledge-based visual question-answering dataset whhere the image content is not sufficient to answer questions. Due to the computational limits, we trained all the models on part of the whole dataset, where the maximum size of the training examples are 80000. Table~\ref{tab:dataset} provides statistics of these dataset.

\paragraph{Base Transformer model}  we utilize the ViLT model with pre-trained weights derived from "BERT-base-uncased". To maintain consistency in our comparisons, we employ the same pre-trained parameters for the ViLT model across all experiments. This configuration entails 11 self-attention blocks (SAB), each with a dimensionality of 768 and 12 attention heads. Additionally, we integrate one task attention block (TAB) following the transformer encoder, which also features 768 hidden dimensions and 12 attention heads.

The model undergoes training on tasks with entirely distinct datasets. Before sending it to the linear projection embedding block, we resize the image from its original dimensions to (384, 640). In the self-attention layers of both the pre-trained transformer and the task-attention block, the image-text feature vector has a shape of (batch, 768).

In contrast to tasks like COCOQA, PathVqa, and SNLI-VE that only require a single input image, NLVR2 necessitates two input images along with a text hypothesis. During the training phase, the text input is merged with one image before being fed into the transformer, and the outputs from the two images are concatenated as one. As a result, the transformer's output becomes (batch, 1536). However, due to the input size constraints of the task-attention block, we condense the vector, $\mathbf{V}$, from 1536 to 768 by computing the average value of adjacent elements:
\begin{equation}
    \textbf{V}'[i] = \frac{\textbf{V}[i] + \textbf{V}[i+1]}{2}, i=0,2,4,...,1534,
\end{equation}
where $\textbf{V}'$ is the compressed feature vector. We then feed $\textbf{V}'$ into the task-attention block.

\paragraph{Baselines for Comparison} since there is no prior method specifically designed for multimodal CL, we use extensions of \textbf{Dytox} \cite{https://doi.org/10.48550/arxiv.2111.11326}, \textbf{FDR} \cite{titsias2020functional}, \textbf{EWC} \cite{kirkpatrick2017overcoming}, \textbf{Experiment Replay} \cite{rolnick2019experience}, and \textbf{Direct Fine-Tuning}, as five alternatives for comparison. Note that direct fine-tuning serves as a lowerbound to measure the effect of catastrophic forgetting and effectiveness of CL. 

For   EWC, we applied a fisher sample percentage of 0.4, meaning that 40\% of the dataset was used to construct the fisher matrix. During training, we set the EWC weight to 0.1.
As for   Experience Replay, the sample size of data in the memory buffer is set at 5\%, and the sample frequency is 100. This choice indicates that for each task, we randomly select 5\% of the data and store it in the memory buffer. During training, after every 100 batches of the current dataset are processed, we randomly pick a batch of data from the memory buffer and train the model on that batch from a randomly chosen previous task.
For TAM-CL, given that the knowledge distillation loss is significantly smaller compared to other losses, we assign a weight of 5000 to $\mathcal{L}_{ikd}$ for all four tasks. Since TAM-CL also employs experience replay as a training strategy, we maintain the same settings for the sample size of the memory (5\%) and the sample frequency (100).

\paragraph{Evaluation Protocol} after learning each task $\mathcal{T}_i$, we assess the forgetting rate on previous tasks $T_k$, where $k$ ranges from 1 to $i-1$. To scrutinize the influence of task order, we conduct experiments with varying task orders that correspond to different levels of task difficulty. For each learned task, we report both the final performance accuracy and the forgetting rates relative to its highest achieved accuracy. Although the agent does not control the task order, it is informative to study the effect of task order on the model performance. To facilitate the analysis of how task order impacts the experiment's performance, we establish an intuitive difficulty ranking for the five tasks based on the metric $\frac{\#\ Training\ Examples}{\#\ Labels}$, as outlined in Table~\ref{tab:dataset}. Essentially, the larger this value, the easier the task is. In this context, we rank the five tasks from most difficult to easiest as follows $\textbf{OKVQA}$, $\textbf{GQA}$, $\textbf{COCOQA}$, $\textbf{SNLI-VE}$, and $\textbf{NLVR2}$, with corresponding scores of 6.19, 43.43, 183.10, 26666.66, and 40000.
To mitigate any potential bias arising from a specific task order, we perform experiments with several different task orders. Additional results can be found in the Appendix.

The majority of Continual Learning (CL) algorithms typically focus on relatively homogeneous tasks. To assess the algorithm's effectiveness in preventing catastrophic forgetting on our diverse set of tasks, we employ a normalization-based metric proposed in \cite{srinivasanclimb}:

\begin{equation}\small
    \mathbb{T}_\textbf{F}(j \leftarrow i) = \frac{S^j_A-S^{j\leftarrow i}_A}{S^j_A-S^j_R},
    \label{eqnor}
\end{equation}
where $\mathbb{T}_\textbf{F}(j \leftarrow i)$ represents the forgetting rate of task $j$ after learning task $i$. $S^j_A$ denotes the accuracy of task $j$ before introducing new tasks, $\mathbb{T}_\textbf{F}(j \leftarrow i)$ represents the accuracy of task $j$ after incorporating task $i$, with $i > j$, and $S^j_R$ signifies the accuracy of task $j$ when labels are randomly selected, computed as $\frac{1}{\#\ labels}$.
Essentially, Eq.~\eqref{eqnor} allows us to compare forgetting rates across tasks that may have significant differences in their nature. This is achieved by evaluating how well the model performs relative to a baseline of total forgetting for that task. 

\paragraph{Optimization Method} for all  tasks, we implement the AdamW optimizer with the following hyperparameters: learning rate $l = 1e-2$, epsilon $\epsilon = 1e-8$, $\beta_1 = 0.9$ and $\beta_2 = 0.98$. This standardized optimization approach ensures consistency and facilitates meaningful comparisons between tasks in our experiments.
Regarding the training epochs, we employ a conservative approach due to computational limitations. Specifically, we conduct training for a limited number of epochs for each task: 5 epochs for SNLI-VE, 10 epochs for COCOQA, GQA, and NLVR2, and 20 epochs for OKVQA. In all our experiments, we utilize a single A100 GPU with a batch size of 16.

\subsection{Comparative Results}
\label{section6.2}

\begin{table*}[h]
%\small
\centering
\begin{tabular}{c|c|c|c|c|c}
    \toprule
    \multicolumn{6}{c}{COCOQA $\rightarrow$ NLVR2 $\rightarrow$ OKVQA $\rightarrow$ SNLI-VE $\rightarrow$ GQA} \\ \midrule
    & COCOQA & NLVR2 & OKVQA & SNLI-VE & GQA \\ \midrule
    \textbf{TAM-CL} & \textbf{66.09} & \textbf{66.07} & \textbf{21.24} & \textbf{64.05} & 50.86 \\ 
    Finetune & 40.67  & 53.85  & 8.26  &  53.83  & \textbf{51.92}\\ 
    FDR & 48.74  & 55.91 & 11.59 & 59.30& 50.67 \\ 
    EWC & 51.44  & 60.87 & 16.16 & 57.93 & 49.67\\ 
    ER & 56.30 & 62.06  & 15.62 & 61.12 & 50.12 \\ 
    Dytox & 60.67& 65.56 & 10.41 & 62.26 & 11.12\\ 
    Avg. & 53.98 & 60.72 & 13.88 & 59.75 & 44.06 \\ \bottomrule
    
\end{tabular}
\caption{Comparison of the accuracy on one task sequences. For each task sequence, each reported value is the \textbf{final accuracy} of that task after learning the last task. The last row represents the average accuracy from different methods on the specific task.}
\label{tab:comparative}
\end{table*}

 In Table \ref{tab:comparative}, we present our performance results after training on all five tasks with the task order of: $\textbf{COCOQA}\rightarrow\textbf{NLVR2}\rightarrow\textbf{OKVQA}\rightarrow\textbf{SNLI-VE}\rightarrow\textbf{GQA}$. As anticipated, fine-tuning exhibits the lowest final accuracy for all tasks except GQA. On the flip side, GQA attains the highest accuracy among all the methods, which aligns with expectations. This observation underscores the inherent trade-off between accuracy and forgetting rate; without any constraints on the forgetting rate, the fine-tuning method is anticipated to achieve relatively higher accuracy at the current task because the learning capacity of the network is free to learn the current task.
 We observe that for most results, the accuracy results of EWC and FDR fall below the average accuracy for all previous tasks. For example, we observe the performance of COCOQA for FDR and EWC are 48.74 and 51.44 respectively, and that for average is 53.98. While EWC and FDR can be effective with smaller models, in line with prior observations for transformers used in learning sequential tasks \cite{srinivasanclimb, jin2021learn, https://doi.org/10.48550/arxiv.2111.11326}, we conclude that regularization-based CL methods are not suitable for CL with large transformers. We deduce that transformers are sensitive to weight consolidation, as their learning capacities are significantly compromised when a large number of weights are frozen. This observation suggests that the learning ability of the transformer networks is distributed among most weights and hence when we use weight consolidation, most of the transformer networks weight are consolidated and as a result the learning ability of the transformer is compromised.
On the contrary, the results for experience replay   demonstrate more promising outcomes. As shown in Table \ref{tab:comparative}, all the accuracy results for experience replay surpass the average line. Such a performance illustrates that replaying the knowledge of previous tasks during the training of the current one can efficiently prevent the parameter distribution shifting away from that of previous tasks, and remind the model of its previous trained results. Although ER is not the best among all the baseline, it still indicates that experience replay proves to be a stable Continual Learning  method, capable of handling models of various sizes and multimodal data. 
We also observe that while Dytox achieves results comparable to our method in some cases, its performance is notably lower in others. For instance, Dytox achieves an accuracy of 10.41 on OKVQA, only higher than fine-tuning and even lower than FDR and EWC. Similarly, the accuracy of Dytox on GQA is only 11.12, which is significantly lower than the rest of methods. Such performance further confirming Dytox's tendency to underfit the current task when trying to balance the accuracy between the previous tasks and the current tasks. Given that Dytox is designed primarily for visual-unimodal tasks, our conclusion is that it is not the most suitable choice for multi-modal tasks.

Comparative experiments  demonstrate that TAM-CL has a superior performance compared to   other alternative method listed in Table \ref{tab:comparative}. Notably, it achieves an 8.93\% higher accuracy than the second best method for COCOQA, a remarkable 31.43\% higher accuracy than the second best method for OKVQA. Moreover, it falls only 2.08\% below the best accuracy for the last task, GQA. This experiment showcases TAM-CL's proficiency in mitigating the forgetting rate of previous tasks while simultaneously upholding a high accuracy on the current task.

\subsection{Ablative Experiments}

We conduct ablation experiments to assess the importance of individual components in our design. We focused on evaluating the impact of the $\mathcal{L}_{ikd}$ loss function, the training strategy involving experience replay, and the utilization of the task attention block. These experiments were conducted using the task sequence order: $\textbf{OKVQA} \rightarrow \textbf{GQA} \rightarrow \textbf{COCOQA} \rightarrow \textbf{SNLI-VE} \rightarrow \textbf{NLVR2}$. 

Results of this experiment are presented in Table \ref{tab:ablative}. To further study the significance of each part, besides the final accuracy, we also present both the forgetting rate of each task which indicates the percentage of dropping between its original accuracy and final accuracy.
Our observation confirms the crucial role of each component in our TAM-CL approach to obtain optimal performance. When comparing the full TAM-CL pipeline to ablation tasks, we consistently observed superior performance in terms of both forgetting rate and accuracy. This result underscores the indispensability of every element in our approach for achieving optimal results. Notably, omitting the $\mathcal{L}_{ikd}$ loss resulted in an average performance drop of 12.74\% across the four tasks, highlighting the critical importance of intermediate knowledge distillation in knowledge transfer and mitigating the forgetting effects.
The observation that ablating the entire task-attention block, which encompasses the $\mathcal{L}{ikd}$ loss, results in slightly better forgetting rates and accuracy performances compared to solely ablating the $\mathcal{L}{ikd}$ loss may initially seem counterintuitive. However, we posit that this discrepancy arises from the fact that the $\mathcal{L}_{ikd}$ loss plays a crucial role in training the task-attention layer, which is the central component for continual learning in the model. As a result, its impact on performance is more pronounced.

\begin{table*}[h]
%\small\centering
\scalebox{0.8}{
\begin{tabular}{c|c|c|c|c|c}
    \toprule
    \multicolumn{6}{c}{OKVQA $\rightarrow$ GQA $\rightarrow$ COCOQA $\rightarrow$ SNLI-VE $\rightarrow$ NLVR2} \\ \midrule
    & OKVQA & GQA & COCOQA & SNLI-VE & NLVR2 \\ \midrule
    TAM-CL  & \textbf{15.09 (54.56\%)} & \textbf{41.91 (27.59\%)} & \textbf{60.86 (18.36\%)} & \textbf{60.47 (29.34\%)} & \textbf{65.86} \\ 
    ablation TAB  & 13.97 (57.22\%) & 39.95 (28.69\%) & 59.92 (19.65\%) & 57.35 (37.22\%) & 65.71\\ 
    ablation $\mathcal{L}_{ikd}$  & 11.53 (65.36\%) & 37.19 (34.07\%) & 58.62 (21.71\%) & 58.34 (35.30\%) & 65.21\\ 
    ablation replay & 2.11 (93.69\%) & 27.25 (51.84\%) & 41.59 (44.33\%) & 50.06 (56.40\%) & 65.64\\ 
    \bottomrule

\end{tabular}
}
\caption{Ablation experiment: the final accuracy ($\uparrow$) and forgetting rate ($\downarrow$) for each task in the ablation methods. }
\label{tab:ablative}
\end{table*}

As anticipated, our training strategy involving experience replay plays an important role in achieving optimal performance by mitigating catastrophic forgetting of previous tasks. We observe that experience replay is particularly beneficial for maintaining the accuracy of tasks early in the sequence. For instance, in Table \ref{tab:ablative}, when experience replay is ablated, the forgetting rate of OKVQA after learning NLVR2 surges to 93.69\%. In contrast, the forgetting rate of GQA after learning NLVR2 is 51.84\%, and the forgetting rate of COCOQA after NLVR2 is 44.33\%, both of which are significantly lower than 93.69\%. This highlights the effectiveness of experience replay in preserving knowledge across tasks.
These results demonstrate that the optimal performance of our method stems from using all the three primary ideas that we proposed.

\subsection{Analytic Experiments}

In this section, we offer analytic experiments to empirically study the various aspects of the proposed algorithm.

\subsubsection{The Effect of the Task Order}

To further analyze the performance of TAM-CL, we compare its performance on four different task sequences and examine the impact of task order on catastrophic forgetting. While we acknowledge that, in practice, we don't have control over the task order as it is determined by the environment, we still aim to understand the effect of task order assuming it is predefined. This analysis provides valuable insights into how TAM-CL adapts to different task sequences.

Tables~4-7 presents the accuracy of current task and the forgetting rate results of previous tasks for each time step, allowing us to evaluate the forgetting rate of previous tasks after the training of every individual task. Upon inspecting the results for these task sequences, we can conclude that the task order plays a crucial role in the performance of CL methods. For instance, in table \ref{tab:table1}, where OKVQA is the third task, its forgetting rate is 22.59\% after learning the final task. On the other hand, in \ref{tab:table2}, where OKVQA is the fourth task, its forgetting rate increases to 27.29\%. Theoretically, as the tasks accumulates, the earlier task will have higher forgetting rate, but the forgetting rate of OKVQA as the thrid task in the first task sequence is lower than that as the fourth task in the second task sequence, which indicates that the task order does affect the performance of each task.

Based on our intuitive metric of task difficulty described in Section \ref{sec:6.1}, we hypothesize that in table \ref{tab:table1}, SNLI-VE and NLVR2 are relatively easier tasks, requiring fewer parameter distribution shifts to achieve high accuracy. As a result, the parameter distribution for OKVQA is less affected. However, in table \ref{tab:table2}, GQA is the second most difficult task, necessitating a larger distribution shift from the previous task, OKVQA, to achieve higher performance. We can observe a similar correlation between forgetting rate and task difficulty in table \ref{tab:table2}. For example, after training on the two relatively easier tasks, NLVR2 and SNLI-VE, the forgetting rates for COCOQA are 7.40\% and 8.45\% respectively, and the forgetting rate increased by only 1.05\%. In contrast, after training on OKVQA and GQA, the forgetting rates for OKVQA are 11.08\% and 19.30\%, which is a 2.63\% and 8.27\% incrasing of forgetting rate, which is significantly higher than after NLVR2 and SNLI-VE. Similarly, in table \ref{tab:table3}, where GQA is the first task, the forgetting rate of GQA after learnt on OKVQA, the third task, is 6.81\%, which is 3.59 higher than its previous task. While SNLI-VE is the fourth task, the forgetting rate of GQA after it is 4.60\%, 2.21\% $\textbf{lower}$ than the forgetting rate of OKVQA, which again proves the relative easiness of SNLI-VE, and thus the hypothesis of section \ref{sec:6.1}. 

We also note an interesting observation: even though NLVR2 is considered a relatively easier task, in table \ref{tab:table1}, after training on NLVR2, the forgetting rates for previous tasks increase by more than twofold. We hypothesize that while NLVR2 may be easy to train, it possesses a fundamental difference from the other tasks. Specifically, NLVR2 takes two images as a single input and performs visual reasoning, whereas all the other tasks only process one image at a time. This unique property of the NLVR2 task leads to a significant shift in parameter distribution, though not to the same extent as the difficulty level of the task.

Furthermore, we have observed an intriguing phenomenon in that after training on certain specific tasks, the forgetting rate of previous tasks can actually decrease. In table \ref{tab:table1}, after training on OKVQA, the forgetting rates for GQA and COCOQA are 6.81\% and 5.81\% respectively. However, after training on the subsequent task, SNLI-VE, the forgetting rates for these tasks decrease to 4.60\% and 4.58\%. At the same time, in table \ref{tab:table4}, the forgetting rate of COCOQA after the second task GQA, is 3.86\%, while that after the third task, SNLI-VE, is 3.37\%. This suggests a potential for forward transfer, which is an aspect we may delve deeper into in future research.

% As discussed in Section 6.2, the two regularization-based methods, EWC and FDR, may not be suitable for large models and long sequence continual learning. We observe that in some cases, the forgetting rate of FDR and EWC is comparable to or even higher than that of   fine-tuning. This observation implies that in such scenarios, these methods may not effectively prevent catastrophic forgetting. However, as all the experiments are done in a single task sequence, it is reasonable to doubt that the specific task order will affect the performance of continual learning methods. To further explore the performance of TAM-CL along with other baseline methods, four more comparative experiments are performed using different task order in the task sequence. 

% For instance, in Table \ref{tab:table1}, after training on OKVQA, the forgetting rate of NLVR2 for the Fine-tuning method is 59.13\%, while for EWC it is 52.07\%, which is close to the performance without any continual learning algorithm. Similarly, after training on GQA, the forgetting rate of OKVQA for the Fine-tuning method is 40.90\% and for EWC it is 42.28\%, which is even higher than the non-CL method baseline. After training on NLVR2, the forgetting rate of SNLI-VE for Fine-tuning is 50.07\%, for FDR it is 47.51\%, and for EWC it is 54.04\%, all of which are close to or above the non-CL baseline.

\begin{table*}[t]
\center
\scalebox{0.8}{
\begin{tabular}{p{1.8cm}|p{2cm}|p{2cm}|p{2cm}|p{2cm}|p{2cm}|p{2cm}}
    \toprule
    \multicolumn{7}{c}{COCOQA $\rightarrow$ NLVR2 $\rightarrow$ OKVQA $\rightarrow$ SNLI-VE $\rightarrow$ GQA} \\ \midrule
    & COCOQA & \multicolumn{2}{|c|}{NLVR2} & \multicolumn{3}{|c}{OKVQA} \\ \midrule
    &  &  & \hfil COCOQA &  &\hfil COCOQA &\hfil NLVR2 \\ \midrule
    TAM-CL &\hfil 76.09 &\hfil \textbf{68.88} &\hfil \textbf{7.89\%} &\hfil 31.63 &\hfil \textbf{9.45\%} &\hfil  \textbf{3.72\%}\\ 
    Finetune &\hfil 75.88 &\hfil 68.71 &\hfil 70.21\% &\hfil \textbf{32.24} & \hfil 19.41\% &\hfil 45.75\%\\ 
    EWC &\hfil 75.89 &\hfil 67.93 &\hfil 73.65\% &\hfil 31.39 &\hfil 20.95\% & \hfil64.20\%\\ 
    FDR &\hfil 71.81 &\hfil 58.43 &\hfil 28.99\% &\hfil 26.01 &\hfil 24.54\% &\hfil 59.07\%\\
    ER &\hfil \textbf{76.77} & \hfil 68.35 & \hfil 12.08\% &\hfil 31.19 &\hfil 13.64\% &\hfil 34.04\%\\
    Dytox &\hfil 76.68 &\hfil 68.41 &\hfil 10.37\% &\hfil 13.93 & \hfil 16.15\% & \hfil 23.35\% \\

\end{tabular}
}
\scalebox{0.8}{
\begin{tabular}{p{1.8cm}|p{3.2cm}|p{3.2cm}|p{3.2cm}|p{3.25cm}}
    \toprule
    & \multicolumn{4}{|c}{SNLI-VE} \\ \midrule
    &  &\hfil COCOQA &\hfil NLVR2 &\hfil OKVQA \\ \midrule
    TAM-CL &\hfil 71.32 &\hfil \textbf{10.25\%} &\hfil \textbf{6.98\%} &\hfil 15.19\% \\ 
    Finetune &\hfil 71.6 &\hfil 31.71\% &\hfil 41.57\% &\hfil 71.71\% \\ 
    EWC &\hfil 71.03 &\hfil 23.55\% &\hfil 43.49\% &\hfil 29.75\% \\ 
    FDR &\hfil 68.56 &\hfil 71.13\% &\hfil 51.00\% &\hfil 52.28\% \\
    ER &\hfil \textbf{71.37} &\hfil 12.05\% &\hfil 10.46\% &\hfil 26.57\% \\
    Dytox &\hfil 69.82 &\hfil 14.30\% &\hfil 16.25\% &\hfil \textbf{2.15\%} \\
    \toprule
    
\end{tabular}
}
\scalebox{0.8}{
\begin{tabular}{p{1.8cm}|p{2.5cm}|p{2.5cm}|p{2.5cm}|p{2.5cm}|p{2.43cm}}
    
    & \multicolumn{5}{|c}{GQA} \\ \midrule
    &  & \hfil COCOQA &\hfil NLVR2 &\hfil OKVQA &\hfil SNLI-VE\\ \midrule
    TAM-CL &\hfil 50.86 &\hfil \textbf{13.15\%} &\hfil \textbf{14.87\%} & \hfil\textbf{22.59\%} &\hfil \textbf{19.13\%} \\ 
    Finetune &\hfil \textbf{51.92} &\hfil 47.09\% & \hfil 79.42\% &\hfil 74.37\% &\hfil 46.81\% \\ 
    EWC &\hfil 49.67 &\hfil 33.10\% &\hfil 42.19\% &\hfil 49.40\% &\hfil 20.45\% \\ 
    FDR &\hfil 50.67 &\hfil 32.12\% &\hfil 29.89\% &\hfil 55.44\% &\hfil 28.26\% \\
    ER & \hfil 50.12 &\hfil 27.20\% & \hfil 34.27\% &\hfil 50.10\% &\hfil 27.19\% \\
    Dytox & \hfil 11.12 &\hfil 20.52\% &\hfil 15.37\% &\hfil 25.26\% &\hfil 20.70\% \\
    \bottomrule
    
\end{tabular}
}
\caption{Accuracy and forgetting rate of task order: COCOQA $\rightarrow$ NLVR2 $\rightarrow$ OKVQA $\rightarrow$ SNLI-VE $\rightarrow$ GQA}
\label{tab:table1}
\end{table*}

\begin{table*}[t]
\center
\scalebox{0.8}{
\begin{tabular}{p{1.8cm}|p{2cm}|p{2cm}|p{2cm}|p{2cm}|p{2cm}|p{2cm}}
    \toprule
    \multicolumn{7}{c}{COCOQA $\rightarrow$ NLVR2 $\rightarrow$ SNLI-VE $\rightarrow$ OKVQA $\rightarrow$ GQA} \\ \midrule
    & COCOQA & \multicolumn{2}{|c|}{NLVR2} & \multicolumn{3}{|c}{SNLI-VE} \\ \midrule
    &  &  & \hfil COCOQA &  &\hfil COCOQA &\hfil NLVR2 \\ \midrule
    TAM-CL & \hfil76.73 & \hfil 69.99 & \hfil\textbf{7.40\%} & \hfil71.12 & \hfil\textbf{8.45\%} &  \hfil\textbf{6.58\%}\\ 
    Finetune & \hfil \textbf{76.95} & \hfil67.83 & \hfil74.75\% & \hfil\textbf{71.72} & \hfil62.51\% & \hfil38.12\%\\ 
    EWC & \hfil76.69 & \hfil \textbf{71.67} & \hfil64.88\% &\hfil 70.84 & \hfil58.61\% & \hfil27.77\%\\ 
    FDR & \hfil72.43 & \hfil57.99 & \hfil31.31\% &\hfil 67.59 & \hfil38.96\% & \hfil65.08\%\\
    ER & \hfil\textbf{76.38} & \hfil70.35 & \hfil8.15\% & \hfil70.69 & \hfil8.65\% & \hfil8.52\%\\
    Dytox & \hfil76.63 &\hfil 68.89 &\hfil 8.50\% &\hfil 70.45 & \hfil9.43\% & \hfil7.94\% \\

\end{tabular}
}
\scalebox{0.8}{
\begin{tabular}
{p{1.8cm}|p{3.2cm}|p{3.2cm}|p{3.2cm}|p{3.25cm}}
    \toprule
    & \multicolumn{4}{|c}{OKVQA} \\ \midrule
    &  & \hfil COCOQA &\hfil NLVR2 & \hfil SNLI-VE \\ \midrule
    TAM-CL & \hfil 30.47 & \hfil \textbf{11.08\%} & \hfil \textbf{14.57\%} & \hfil 19.23\% \\ 
    Finetune & \hfil \textbf{32.12} & \hfil 19.19\% & \hfil 59.13\% & \hfil 22.22\% \\ 
    EWC & \hfil 31.68 & \hfil 18.83\% & \hfil 52.07\% & \hfil \textbf{15.44\%} \\ 
    FDR & \hfil 11.49& \hfil 44.23\% & \hfil 45.52\%& \hfil 41.27\% \\
    ER & \hfil 30.91 & \hfil 12.92\% & \hfil 28.12\% & \hfil 17.92\% \\
    Dytox & \hfil 19.12 & \hfil 17.16\% & \hfil 32.55\% & 
 \hfil 37.06\% \\

\end{tabular}
}
\scalebox{0.8}{
\begin{tabular}{p{1.8cm}|p{2.5cm}|p{2.5cm}|p{2.5cm}|p{2.5cm}|p{2.43cm}}
    \toprule
    & \multicolumn{5}{|c}{GQA} \\ \midrule
    &  & \hfil COCOQA & \hfil NLVR2 &\hfil SNLI-VE &\hfil OKVQA\\ \midrule
    TAM-CL & \hfil 56.20 & \hfil \textbf{19.30\%} & \hfil \textbf{23.34\%} & \hfil \textbf{24.98\%} & \hfil \textbf{27.29\%} \\ 
    Finetune & \hfil 57.03 & \hfil 36.53\% & \hfil 66.16\% & \hfil 34.11\% & \hfil 40.90\% \\ 
    EWC & \hfil \textbf{57.21} & \hfil 33.51\% & \hfil 60.60\% & \hfil 32.84\% & \hfil 42.28\% \\ 
    FDR & \hfil 42.67 & \hfil 41.08\% & \hfil 46.41\%& \hfil 74.08\% & \hfil 50.17\%\\
    ER & \hfil 57.04 & \hfil 26.04\% & \hfil 32.89\% & \hfil 27.97\% & \hfil 41.33\% \\
    Dytox & \hfil 6.10 & \hfil 21.81\% & \hfil 38.50\% & \hfil 30.10\% & \hfil 34.43\% \\
    \toprule
    
\end{tabular}
}
\caption{Accuracy and forgetting rate of task order: COCOQA $\rightarrow$ NLVR2 $\rightarrow$ SNLI-VE $\rightarrow$ OKVQA $\rightarrow$ GQA}
\label{tab:table2}
\end{table*}

\begin{table*}[t]
\center
\scalebox{0.8}{
\begin{tabular}{p{1.8cm}|p{2cm}|p{2cm}|p{2cm}|p{2cm}|p{2cm}|p{2cm}}
    \toprule
    \multicolumn{7}{c}{GQA $\rightarrow$ COCOQA $\rightarrow$ OKVQA $\rightarrow$ SNLI-VE $\rightarrow$ NLVR2} \\ \midrule
    &GQA & \multicolumn{2}{|c|}{COCOQA} & \multicolumn{3}{|c}{OKVQA} \\ \midrule
    &  &  & \hfil GQA &  &\hfil GQA &\hfil COCOQA\\ \midrule
    TAM-CL &\hfil 56.58 &\hfil 74.44 &\hfil 3.22\% &\hfil 30.44 &\hfil \textbf{6.81\%} &\hfil 5.81\%\\ 
    Finetune &\hfil 58.11 &\hfil 75.95 &\hfil 10.60\% &\hfil \textbf{31.99} & \hfil 12.78\% &\hfil 10.65\%\\ 
    EWC &\hfil \textbf{58.69} &\hfil \textbf{76.12} &\hfil 10.45\% &\hfil 31.15 &\hfil 13.46\% & \hfil 7.97\%\\ 
    FDR &\hfil 55.50 &\hfil 71.42 &\hfil 12.70\% &\hfil 28.39 &\hfil 12.17\% &\hfil 6.21\%\\
    ER &\hfil 57.61 & \hfil 75.17 & \hfil 4.17\% &\hfil 31.21 &\hfil 9.58\% &\hfil \textbf{4.26\%}\\
    Dytox &\hfil 56.83 &\hfil 75.22 &\hfil \textbf{1.93\%} &\hfil 29.87 & \hfil 11.29\% & \hfil 7.87\% \\

\end{tabular}
}
\scalebox{0.8}{
\begin{tabular}{p{1.8cm}|p{3.2cm}|p{3.2cm}|p{3.2cm}|p{3.25cm}}
    \toprule
    & \multicolumn{4}{|c}{SNLI-VE} \\ \midrule
    &  &\hfil GQA &\hfil COCOQA &\hfil OKVQA \\ \midrule
    TAM-CL &\hfil 71.49 &\hfil \textbf{4.60\%} &\hfil \textbf{4.58\%} &\hfil \textbf{8.51\%} \\ 
    Finetune &\hfil \textbf{72.54} &\hfil 9.91\% &\hfil 7.35\% &\hfil 15.93\% \\ 
    EWC &\hfil 72.10 &\hfil 10.82\% &\hfil 6.43\% &\hfil 11.46\% \\ 
    FDR &\hfil 70.17 &\hfil 8.29\% &\hfil 7.11\% &\hfil 13.43\% \\
    ER &\hfil 72.39 &\hfil 6.00\% &\hfil 5.03\% &\hfil 10.47\% \\
    Dytox &\hfil 72.44 &\hfil 8.13\% &\hfil 5.40\% &\hfil 10.07\% \\
    \toprule
    
\end{tabular}
}
\scalebox{0.8}{
\begin{tabular}{p{1.8cm}|p{2.5cm}|p{2.5cm}|p{2.5cm}|p{2.5cm}|p{2.43cm}}
    
    & \multicolumn{5}{|c}{NLVR2} \\ \midrule
    &  & \hfil GQA &\hfil COCOQA &\hfil OKVQA &\hfil SNLI-VE\\ \midrule
    TAM-CL &\hfil 70.04 &\hfil \textbf{9.43\%} &\hfil \textbf{9.72\%} & \hfil\textbf{20.72\%} &\hfil \textbf{14.98\%} \\ 
    Finetune &\hfil 70.32 &\hfil 55.09\% & \hfil 63.90\% &\hfil 77.80\% &\hfil 53.18\% \\ 
    EWC &\hfil 67.57 &\hfil 57.47\% &\hfil 63.77\% &\hfil 80.89\% &\hfil 51.90\% \\ 
    FDR &\hfil 53.60 &\hfil 33.54\% &\hfil 56.20\% &\hfil 63.64\% &\hfil 36.42\% \\
    ER & \hfil \textbf{70.38} &\hfil 13.19\% & \hfil 9.90\% &\hfil 23.43\% &\hfil 19.93\% \\
    Dytox & \hfil 67.34 &\hfil 15.86\% &\hfil 13.90\% &\hfil 41.45\% &\hfil 27.61\% \\
    \bottomrule
    
\end{tabular}
}
\caption{Accuracy and forgetting rate of task order: GQA $\rightarrow$ COCOQA $\rightarrow$ OKVQA $\rightarrow$ SNLI-VE $\rightarrow$ NLVR2}
\label{tab:table3}
\end{table*}

\begin{table*}[t]
\center
\scalebox{0.8}{
\begin{tabular}{p{1.8cm}|p{2cm}|p{2cm}|p{2cm}|p{2cm}|p{2cm}|p{2cm}}
    \toprule
    \multicolumn{7}{c}{COCOQA $\rightarrow$ GQA $\rightarrow$ SNLI-VE $\rightarrow$ OKVQA $\rightarrow$ NLVR2} \\ \midrule
    & COCOQA & \multicolumn{2}{|c|}{GQA} & \multicolumn{3}{|c}{SNLI-VE} \\ \midrule
    &  &  & \hfil COCOQA &  &\hfil COCOQA &\hfil GQA\\ \midrule
    TAM-CL &\hfil 76.59 &\hfil 57.83 &\hfil \textbf{3.86\%} &\hfil 72.03 &\hfil \textbf{3.37\%} &\hfil  \textbf{1.66\%}\\ 
    Finetune &\hfil 76.84 &\hfil \textbf{58.72} &\hfil 18.86\% &\hfil 72.09 & \hfil 23.57\% &\hfil 5.64\%\\ 
    EWC &\hfil 76.49 &\hfil 57.27 &\hfil 18.65\% &\hfil 72.11 &\hfil 21.18\% & \hfil 4.26\%\\ 
    FDR &\hfil 72.00 &\hfil 53.42 &\hfil 14.66\% &\hfil 68.25 &\hfil 20.47\% &\hfil 2.74\%\\
    ER &\hfil \textbf{76.88} & \hfil 58.07 & \hfil 17.52\% &\hfil 72.19 &\hfil 9.01\% &\hfil 1.71\%\\
    Dytox &\hfil 76.77 &\hfil 25.41 &\hfil 4.51\% &\hfil \textbf{72.33} & \hfil 3.77\% & \hfil 2.14\% \\

\end{tabular}
}
\scalebox{0.8}{
\begin{tabular}{p{1.8cm}|p{3.2cm}|p{3.2cm}|p{3.2cm}|p{3.25cm}}
    \toprule
    & \multicolumn{4}{|c}{OKVQA} \\ \midrule
    &  &\hfil COCOQA &\hfil GQA &\hfil SNLI-VE \\ \midrule
    TAM-CL &\hfil 30.74 &\hfil \textbf{6.64\%} &\hfil \textbf{4.78\%} &\hfil 5.66\% \\ 
    Finetune &\hfil 32.77 &\hfil 24.37\% &\hfil 9.89\% &\hfil 9.27\% \\ 
    EWC &\hfil 31.78 &\hfil 22.88\% &\hfil 5.88\% &\hfil 8.65\% \\ 
    FDR &\hfil 26.57 &\hfil 26.88\% &\hfil 8.84\% &\hfil \textbf{4.77\%} \\
    ER &\hfil \textbf{33.22} &\hfil 11.64\% &\hfil 6.11\% &\hfil 5.74\% \\
    Dytox &\hfil 13.74 &\hfil 10.61\% &\hfil 4.98\% &\hfil 6.74\% \\

\end{tabular}
}
\scalebox{0.8}{
\begin{tabular}{p{1.8cm}|p{2.5cm}|p{2.5cm}|p{2.5cm}|p{2.5cm}|p{2.43cm}}
    \toprule
    & \multicolumn{5}{|c}{NLVR2} \\ \midrule
    &  & \hfil COCOQA &\hfil GQA &\hfil SNLI-VE &\hfil OKVQA\\ \midrule
    TAM-CL &\hfil \textbf{69.15} &\hfil \textbf{10.09\%} &\hfil \textbf{11.74\%} & \hfil\textbf{14.60\%} &\hfil \textbf{16.79\%} \\ 
    Finetune &\hfil 67.27 &\hfil 76.18\% & \hfil 55.89\% &\hfil 50.07\% &\hfil 75.11\% \\ 
    EWC &\hfil 68.43 &\hfil 78.72\% &\hfil 54.84\% &\hfil 54.04\% &\hfil 79.46\% \\ 
    FDR &\hfil 58.73 &\hfil 38.38\% &\hfil 40.97\% &\hfil 47.51\% &\hfil 52.12\% \\
    ER & \hfil 67.63 &\hfil 16.08\% & \hfil 12.26\% &\hfil 24.12\% &\hfil 19.80\% \\
    Dytox & \hfil 69.01 &\hfil 18.20\% &\hfil 23.63\% &\hfil 26.89\% &\hfil 29.90\% \\
    \bottomrule
    
\end{tabular}
}
\caption{Accuracy and forgetting rate of task order: COCOQA $\rightarrow$ GQA $\rightarrow$ SNLI-VE $\rightarrow$ OKVQA $\rightarrow$ NLVR2}
\label{tab:table4}
\end{table*}

Overall, we also observe that   experience replay has a relatively high accuracy and low forgetting rates compared to the two regularization-based methods. However, its capacity to prevent catastrophic forgetting in later tasks is not as consistent as in early tasks. For example, in Table \ref{tab:table1}, after training on the last task, GQA, the forgetting rate of OKVQA for ER is 50.10\%, which is close to FDR and higher than EWC. In contrast, after training on SNLI-VE, the forgetting rate of OKVQA for ER is 26.57\%, considerably better than the forgetting rate for FDR, which is 52.28\%.
On the other hand,   Dytox   exhibits stable capacity to prevent catastrophic forgetting, which is not significantly affected by the length of the task sequence. However, it loses competitiveness due to its lower performance in training the current task. For example, in Table \ref{tab:table1}, the accuracy of OKVQA for Dytox is 13.93, while the accuracy for all the other methods is above 25. Additionally, Dytox only achieves 11.12\% in GQA, whereas the accuracy for all other methods is above 50\%. In table \ref{tab:table4}, the accuracy of Dytox on OKVQA is only 19.12 and that on GQA is even 6.10, which are both significantly lower than the performance of other methods.
Ultimately, although TAM-CL may not always lead in every individual forgetting rate, it surpasses the other methods in 87.5\% of the total forgetting rates. In the remaining exceptional cases, TAM-CL still demonstrates above-average performance, underscoring its capacity and stability in preventing catastrophic forgetting.
Note that while the accuracy of TAM-CL may not always be the highest among the six methods, the differences between the top accuracy and TAM-CL's accuracy are consistently below 5\%. Given that our main focus is on the improvement of forgetting rate, we consider these slight accuracy differences to be within an acceptable range.

\subsubsection{Fine-tuning on Unimodal Tasks}

Despite focusing on multimodal learning, the ability for unimodal learning is important. The reason is that in practice, some of the modalities maybe occluded. For this reason, we performed experiments on using our approach on unimodal tasks.
We tested our method's performance on three different unimodal vision datasets, coco-classification \cite{lin2014microsoft}, inaturalist \cite{vanhorn2018inaturalist} and places365 \cite{zhou2017places}, along with other baseline performance shown in \ref{section6.2}.To apply the multimodal transformer backbone, ViLT, to unimodal vision-only task, we created a vision-language pair for every image data, while the language input is universally ``This is an image'' for all the images.

We also performed experiments on three different natural language datasets, including PIQA \cite{bisk2020piqa}, CommonsenseQA \cite{talmor2018commonsenseqa}, and Hellaswag \cite{zellers2019hellaswag}. To form image-text pairs for the multimodal transformer input, an averaged image of all MS-COCO training image dataset is used to pair with every text sample in the dataset. We note that as ViLT only allows language token with maximum size of 40 as input which might not be sufficient for language tasks. To deal with this issue,   we first down-sample the meaningless image and thus reduce its token length from 144 to 16. Then, we extend the available length token by creating copies of pre-trained ViLT's language positional embeddings, and concatenate the embeddings to get the extended positional embeddings.

In table \ref{table:unimodal}, we present our experimental results for vision-only and language-only unimodal tasks. We report the final accuracy after training the model  on the last task, PLACES365. As shown in the table, although TAM-CL is specifically designed for multimodal continual learning setting, it's performance on vision-only tasks on average remains the top performer. For COCO classification and iNaturalist datasets, TAM-CL achieves 73.03\% and 50.63\% higher than all the baseline scores, respectively. Regarding the accuracy of PLACES365, EWC achieves the best score, 43.64\%, and finetuning get the second high performance. Although that of TAM-CL is not the best performer, there's only a 0.4\% difference between TAM-CL's accuracy and the best performer. Overall, TAM-CL achieves the top performance in vision-only task compared with those unimodal continual learning methods.

\begin{table*}
\centering
\footnotesize
\scalebox{1}{
\begin{tabular}{c|c|c|c|c}
    \toprule
     \multicolumn{5}{c}{COCO$\rightarrow$iNaturalist$\rightarrow$PLACES365} \\ \midrule
    & COCO & \hfil iNaturalist & \hfil PLACES365 & \hfil Avg.\\ \midrule
    \textbf{TAM-CL} & \hfil \textbf{73.03} & \hfil \textbf{50.63} & \hfil 43.44 & \hfil \textbf{55.66} \\ 
    Dytox & \hfil 72.56 & \hfil 48.88 & \hfil 43.29 & \hfil 54.91 \\ 
    Replay & \hfil 71.91 & \hfil 48.79 & \hfil 43.18 & \hfil 54.62 \\ 
    EWC & \hfil 62.22 & \hfil 48.71 & \hfil \textbf{43.64} & \hfil 51.52\\
    Finetune & \hfil 62.21 & \hfil 47.56 & \hfil 43.58 & \hfil 51.11\\
    \toprule
    \multicolumn{5}{c}{COMMONSENSEQA$\rightarrow$PIQA$\rightarrow$HELLASWAG} \\ \midrule
    & COMMON & \hfil PIQA & \hfil HELLASWAG & \hfil Avg. 
    \\ \midrule
    TAM-CL & \hfil \textbf{21.43} & \hfil \textbf{52.45} & \hfil 24.98  & \hfil \textbf{32.95}  \\ 
    Dytox & \hfil 21.21 & \hfil 47.99 & \hfil \textbf{25.29} & \hfil 31.49 \\ 
    Replay & \hfil 20.25 & \hfil 51.89 & \hfil 24.87 & \hfil 32.30 \\ 
    EWC & \hfil 21.38 & \hfil 50.33 & \hfil 25.10 & \hfil 32.27 \\
    Finetune & \hfil 21.38 & \hfil 49.62 & \hfil 24.99 & \hfil 31.99 \\
    \toprule
    
\end{tabular}
}
\caption{Comparison Experiment with Unimodal Tasks. (Upper) The accuracy of different methods on vision-only task sequence after trained on the last task. (Lower) The accuracy of different methods on language-only task sequence after trained on the last task. }
\label{table:unimodal}
\end{table*}

As for the language-only unimodal tasks, TAM-CL outperforms other baseline methods with respect to the non-last tasks. It achieves 21.43\% accuracy on COMMONSENSEQA and 52.45\%  on PIQA, after trained on HELLASWAG dataset. Dytox, on the other hand, has the worst performance on PIQA, which is 47.99\%. As Dytox is designed specifically for vision-only task, its underperformance on language-only tasks is not something unexpected. However, it achieves the best score on the last task, 25.29\%  on HELLASWAG, which indicates its capacity of learning the current task, although it cannot efficiently prevent the catastrophic forgetting of previous tasks. 
We conclude that although TAM-CL does not achieve the same level of performance  on unimodal tasks, it still generally outperforms many SOTA continual learning methods on unimodal tasks. These results offers an advantage for TAM-CL in that it can perform relatively well in situations in which one of the inputs is occluded.

\subsubsection{Hyper-parameter Sensitivity Analysis}

We study the sensitivity of the TAM-CL with respect to important hyperparameters of our approach.
To this end, we compared the performance of our proposed method  with variant length of the task token, $\tau$, and the number of frozen layers of the backbone transformer's encoder.

\begin{table*}
\centering
\footnotesize
\scalebox{1.2}{
\begin{tabular}{c|c|c|c|c|c}
    \toprule
     \multicolumn{6}{c}{COCOQA$\rightarrow$NLVR2$\rightarrow$SNLI-VE$\rightarrow$OKVQA$\rightarrow$GQA} \\ \midrule
    & GQA & \hfil COCOQA & \hfil NLVR2 &\hfil SNLI-VE &\hfil OKVQA\\ \midrule
    $\tau \ \times$ 1 & \hfil 51.23 & \hfil 64.64 & \hfil \textbf{65.60} & \hfil \textbf{21.03} & \hfil 65.84 \\ 
    $\tau \ \times$ 3 & \hfil 50.89 & \hfil 64.94 & \hfil 61.95 & \hfil 19.83 & \hfil \textbf{67.33} \\ 
    $\tau \ \times$ 5 & \hfil \textbf{51.66} & \hfil 64.85 & \hfil 61.95 & \hfil 19.47 & \hfil 65.34 \\ 
    $\tau \ \times$ 10 & \hfil 51.25 & \hfil \textbf{66.77} & \hfil 63.09& \hfil 20.31 & \hfil 66.30\\
    \midrule
    Freeze 0 & \hfil 49.15 & \hfil 58.18 & \hfil 62.37  & \hfil 21.86 & \hfil 66.00 \\ 
    Freeze 3 & \hfil 50.47 & \hfil 63.14 & \hfil 62.25 & \hfil \textbf{22.48} & \hfil \textbf{67.20} \\ 
    Freeze 6 & \hfil \textbf{51.23} & \hfil \textbf{65.78} & \hfil \textbf{65.60} & \hfil 21.03 & \hfil 65.84 \\ 
    Freeze 9 & \hfil 50.57 & \hfil 65.48 & \hfil 63.06& \hfil 17.92 & \hfil 62.50\\
    Freeze all & \hfil 49.79 & \hfil 60.81 & \hfil 61.46& \hfil 15.98 & \hfil 65.36\\
    \toprule
    
\end{tabular}
}
\caption{Experiments of variant of task token length and number of frozen layers on one multimodal task sequence. The accuracy for each task after trained on the last task, GQA, is presented. (Upper) The accuracy of different token length $\tau \times$ N, where N is the integer multiplier. (Lower) The accuracy of freezing different number of encoder layers. }
\label{table:hyperparameter}
\end{table*}

As task token is a key component that carries the task-specific knowledge and help preserve the performance of previous tasks during the continual learning, we want to analyze whether the change of task token size will affect its capacity to prevent catastrophic forgetting. With all the other hyper-parameter fixed, we run   experiments on varying dimensions for task token, $\tau$, with size of 768, 2304, 3840 and 7680, respectively. In Table \ref{table:hyperparameter}, we observe that the accuracy gap between different $\tau$ size does exist, but none of the $\tau$ achieves the absolutely leading performance. The default setting, 768, gets the best score on the accuracy of NLVR2 and SNLI-VE, which are 65.60 and 21.03 respectively, but under-performed by  GQA, COCOQA and OKVQA compared with other $\tau$ sizes. At the same time, with $\tau$ size of 2304, which is three times larger than the default, TAM-CL achieve the best accuracy on OKVQA, 67.33. With $\tau$ size of 3840, TAM-CL achieve the best on the last task, the GQA dataset. With 10 size of $\tau$, TAM-CL outperforms all the other on COCOQA, 66.77. Throughout all the experiment results, besides the performance of default $\tau$ on NLVR2, which outperforms the second best by 3.97\%, the leading accuracy of specific $\tau$ size on different tasks relatively trivial, which indicates that the variant of task token $\tau$ does not perform the key rule in affecting the performance of TAM-CL as a whole. This conclusion is important because it shows sensitivity with respect to this hyper-parameter is not significant and we do not need to fine-tune it extensively for optimal performance.

We also studied the effect of  freezing different number of layers of the transformer encoder during the training stage on the downstream performance. We observe that by freezing 0 and 3 layers, the performance on SNLI-VE and OKVQA would be the best and the second best scores, respectively.
Our default setting, which freezes 6 layers, outperforms all other setting on GQA, COCOQA and NLVR2. Nevertheless, by freezing 9 and all encoder layers, the performance of SNLI-VE drops significantly, 17.92\% and 15.98\% compared with  all above 21.00\%. We speculate that   less frozen layers in the encoder, the  model can learn  a single task better. However, its ability of preventing catastrophic forgetting would decrease. Meanwhile, if most of the encoder layers are frozen, it harms the capacity of the model to learn the current task. Also, the preservation of performance on previous tasks is not improved. We conclude that freezing 6 layers is an optimal choice which on average leads to acceptable performance on most datasets.

\section{Conclusion}
We developed a multimodal continual learning algorithm when a transformer architecture is the base model. My approach involves dynamic model expansion, through task-specific attention, knowledge distillation, and experience replay to mitigate catastrophic forgetting and enabling positive knowledge transfer. The experiments demonstrate the effectiveness of your approach, achieving state-of-the-art performance in terms of both forward transfer and catastrophic forgetting. The TAM-CL architecture opens up new avenues for studying continual learning in multimodal settings. Future works include exploring using adapter weights in architecture to improve the learning efficiency during the training phase.

\bibliography{ref}
\bibliographystyle{plain}

\clearpage



% \bibitem[ Wang, Zhen and Liu, Liu and Duan, Yiqun and Kong, Yajing and Tao, Dacheng(2022)]{wang2022continual}

% \end{thebibliography}
\end{document}